\documentclass[a4paper,onecolumn]{article}

\usepackage{graphicx}
\usepackage{amssymb,amsmath,bm}
\usepackage{textcomp}
\usepackage{enumitem}
\usepackage[margin=1.2in]{geometry}
\usepackage{authblk}

\setlist[itemize]{leftmargin=*}

\newcommand{\si}{{s_i}}

\newcommand{\cij}{{c_{ji}}}

\title{Scoring Formulation for Multi-Condition Joint PLDA}

\author[1]{Luciana Ferrer}
\affil[1]{\mbox{Instituto de Investigaci\'on en Ciencias de la Computaci\'on, }
\mbox{CONICET-Universidad de Buenos Aires, Buenos Aires, Argentina}}
\date{}
\begin{document}
  \maketitle

The joint PLDA model, originally proposed in \cite{ferrer2017joint} and then further developed and tested for a multilingual speaker recognition task in \cite{ferrer:jmlr18}, is a generalization of PLDA where the nuisance variable is no longer considered independent across samples, but potentially shared (tied) across samples that correspond to the same nuisance condition. The original work considered a single nuisance condition, deriving the EM and scoring formulas for this scenario. In this document, we show how to obtain likelihood ratios for scoring when multiple nuisance conditions are allowed in the model. 

We assume that the within-speaker variability can be decomposed into $N$ terms corresponding to different nuisance conditions that could correspond to, for example, the language spoken in the sample, the microphone type, the noise type and level, or any other characteristic of the sample that can be considered to occur independently of all other characteristics, plus the usual noise term. That is, we propose to model vector $m_i$ for sample $i$ as:
\begin{equation}
m_i = \mu + V y_\si + \sum_j U_j x^j_\cij + \epsilon_i,
\end{equation}
where $y_\si$ is a vector of size $R_y$, $j$ is an index over the nuisance conditions, $x^j_\cij$ is a vector of size $R_{xj}$, and
\begin{eqnarray}
y_s & \sim & N(0,I), \forall s\\
x^j_{c_j} & \sim & N(0,I), \forall j, c_j \\
\epsilon_i & \sim & N(0, D^{-1}).
\end{eqnarray}
where all variables are assumed independent of each other.

For the purpose of the derivations below, we assume that the global mean on the training data has already been subtracted from the vectors $m_i$, so that $\mu=0$. 

The model's parameters to estimate are $\lambda = \{\mu, V, U_1, \ldots, U_N, D\}$. The input data for the training algorithm is now required to have the nuisance condition labels for each sample as well as the speaker labels, as usual.

Let us define:
\begin{eqnarray}
z_i & = & \begin{bmatrix}y_\si^T & (x^1_{c_{1i}})^T & \ldots & (x^N_{c_{Ni}})^T \end{bmatrix}^T\label{eq:z} \\
W & = & \begin{bmatrix}V U_1 \ldots U_N\end{bmatrix}\label{eq:w}
\end{eqnarray}
so that
\begin{equation}
m_i = \mu + W z_i + \epsilon_i,
\end{equation}
Here, $z_i$ has dimension $R_z = R_y+\sum_j R_{xj}$.

In the following, we take
\begin{eqnarray}
Y & = & \{y_1, \ldots, y_S \} \\
X^j & = & \{x^j_1, \ldots, x^j_{C_j} \}, \forall j=1,\ldots,N \\
Z & = & \{z_1, \ldots, z_I \} \\
M & = & \{m_1, \ldots, m_I \} 
%X & = & \{X^1, \ldots, X^N \}
\end{eqnarray}
where $S$ is the total number of speakers, $C_j$ is the total number of labels for condition $j$, and $I$ is the number of samples.

\section{Prior}

The joint prior for the hidden variables for all the data is given by
\begin{eqnarray}
\label{eq:prior}
p(Y,X^1,\ldots,X^N) & = & p(Y) \prod_j p(X^j) \\
& \propto & \exp(-\frac{1}{2} \sum_s y^T_s y_s -\frac{1}{2} \sum_j \sum_{c_j} (x^j_{c_j})^T x^j_{c_j}) \\
& = & \exp(-\frac{1}{2} \sum_i z_i^T P_i z_i)
\end{eqnarray}
where
\begin{equation}
P_i = \text{diag}\left(\frac{1}{n_\si}I,\frac{1}{n_{c_{1i}}}I,\ldots,\frac{1}{n_{c_{Ni}}}I\right)
\end{equation}
where $n_\cij$ is the number of samples with label $\cij$ (i.e., the label for condition $j$ for sample $i$) and $n_\si$ is the number of samples for speaker $\si$. The function diag creates a block diagonal matrix with all the matrices in its input. In this case, the resulting $P_i$ matrix is diagonal, since all input matrices are also diagonal.

\section{Likelihood}

The full data likelihood is given by

\begin{eqnarray}
\label{eq:llk} 
p(M|Z,\lambda) & = & \prod_i N(m_i|Wz_i,D^{-1}) \nonumber \\
& \propto &   \exp  \sum_i \left(-\frac{1}{2} (m_i - Wz_i)^T D (m_i - Wz_i) + \frac{1}{2} \log|D| \right) \nonumber \\
 & = & \exp \sum_i \left( -\frac{1}{2} m^T_i D m_i + m^T_i D W z_i -\frac{1}{2}  z_i^T W^T D W z_i + \frac{1}{2} \log|D|         \right) 
\end{eqnarray}

\section{Joint}

The joint probability is given by the product of the likelihood and the prior. Keeping only the terms that depend on either $M$ or $Z$, we get
\begin{eqnarray}
p(M,Z|\lambda) & \propto & \exp \sum_i \left( -\frac{1}{2} m^T_i D m_i + m^T_i D W z_i -\frac{1}{2}  z^T_i W^T D W z_i -\frac{1}{2} z_i^T P_i z_i  \right) \\
 & = & \exp \sum_i \left( -\frac{1}{2} m^T_i D m_i + m^T_i D W z_i - \frac{1}{2}z^T_i K_i z_i \right) 
\end{eqnarray}
where 
\begin{eqnarray}
K_i & = & W^T D W + P_i 
\end{eqnarray}

\section{Posterior}

The posterior probability is proportional to the joint probability. Keeping only the terms that depend on $Z$ we get
\begin{equation}
p(Z|M,\lambda) \propto \exp \sum_i \left( m^T_i D W z_i - \frac{1}{2} z^T_i K_i z_i \right)
\end{equation}

\section{Scoring for single-enrollment and \\  single-test trials}

As for standard PLDA, we define the score as the likelihood ratio between the two hypotheses: that the speakers are the same and that the speakers are different. Nevertheless, in this case we need to marginalize both likelihoods over new hypotheses: that each nuisance condition is the same or different in the two samples. Hence, the LR for single-enrollment and single-test trials is computed as follows:
\begin{eqnarray}
LR & = & \frac{\sum_{h \in \cal H} p(m_E,m_T|H_{SS},h)P(h|H_{SS})}{\sum_{h \in \cal H} p(m_E,m_T|H_{DS},h)P(h|H_{DS})}\nonumber
\end{eqnarray}
where $H_{SS}$ is the hypothesis that the speakers for both sets are the same,  $H_{DS}$ is the hypothesis that they are different, and ${\cal H}$ is the set of all possible combinations of hypothesis about the nuisance conditions 
\begin{eqnarray}
{\cal H}= \{(H_{l_1C_1},\ldots,H_{l_NC_N}): l_j \in \{S,D\}, \forall j \in \{1,\ldots,N\}\} 
\end{eqnarray}
where $S$ and $D$ refer to the hypotheses that the conditions are the same and different, respectively. 

The priors $P(h|H_{SS})$ and $P(h|H_{DS})$ can be set to any probability distribution, depending on the task, as they are parameters of the method. For our experiments, we assume independence across priors for the different conditions so that $P(h|H_{SS}) = \prod_j P(H_{m_jC_j}|H_{SS})$ and, similarly, for $P(h|H_{DS})$, where each individual prior can be set independently. 

Define $M=\{m_E,m_T\}$ and $h_a=\{H_{l_SS},h\}$ (where $l_S \in \{S,D\}$ indicates the hypothesis about the speaker identity) to be the overall hypothesis including the hypotheses about the speaker and all conditions. We can now write:
\begin{eqnarray}
\label{eq:likM}
p(m_E,m_T|H_{l_SS},h) = p(M|h_a) = \left. \frac{p(M|Z,h_a)  p(Z|h_a)}{p(Z | M, h_a) } \right|_{Z=0}
\end{eqnarray}
Here, $Z$ is the concatenation of all the latent variables for both samples. For each case (condition or speaker) there will be two separate variables if the hypothesis for that variable is that they are different, or a single variable if the hypothesis is that they are the same.
In the following we assume that the hypotheses are such that first M variables are tied and the last N+1-M are not tied. We can then split $Z$ in three parts:
\begin{equation}
Z^T = [Z_S^T Z_E^T Z_T^T]
\end{equation}
where $Z_S$ has size $N_S$ and refers to the concatenation of the latent variables that are shared for both enrollment and test samples (i.e., the latent variables corresponding to the conditions that are hypothesized to be the same for both samples) and $Z_E$ and $Z_T$, each of size $N_D$, refer to the concatenation of the latent variables that are not shared, for each of the two samples. 

Now, the likelihood in the numerator of Equation (\ref{eq:likM}) is the same for all hypotheses since, regardless of whether the latent variables are tied or not, Equation (\ref{eq:llk}) has the same form. Hence, that term cancels out in the computation of the LR. So, to compute the LR we just need to find the expression for $p(Z = 0| M, h_a)$ for all possible hypotheses $h_a$.
To this end, we write the posterior as a function of the three terms of $Z$ above:
\begin{eqnarray}
p(Z | M, h_a) % & \propto & \exp  \left( m_E^T D W [Z_S^T Z_E^T]^T - \frac{1}{2} [Z_S^T Z_E^T] W^T D W  [Z_S^T Z_E^T]^T + \right. \nonumber \\
%& & \left. m_T^T D W [Z_S^T Z_T^T]^T - \frac{1}{2} [Z_S^T Z_T^T] W^T D W  [Z_S^T Z_T^T]^T \right.\nonumber \\
%& & \left. -\frac{1}{2} (Z_S^T Z_S + Z_E^T Z_E + Z_T^T Z_T) \right) \\
& \propto & \exp  \left( m_E^T D W [Z_S^T Z_E^T]^T - \frac{1}{2}[Z_S^T Z_E^T] K  [Z_S^T Z_E^T]^T + \right. \nonumber \\
& & \left. m_T^T D W [Z_S^T Z_T^T]^T - \frac{1}{2} [Z_S^T Z_T^T] K [Z_S^T Z_T^T]^T\right)
\end{eqnarray}
where
\begin{eqnarray}
K & = & W^T D W + P \\
W & = & [W_S W_D]
\end{eqnarray}
where $W_S$ and $W_D$ are formed by concatenating the matrices $V$ or $U_j$ corresponding to the variables assumed to be the same or different, respectively, in the same order used to form $Z_S$ and $Z_E$ or $Z_T$, and
\begin{eqnarray}
\label{eq:Psc}
P & = & \text{diag}(\frac{1}{2}I_{N_S}, I_{N_D}) 
\end{eqnarray}
where the subindex of $I$ indicates the dimension. 

Now, putting the posterior as a function of $Z$ to find the form of its distribution,
\begin{eqnarray}
p(Z | M, h_a) & \propto & \exp  \left( (m_E^T D W_E + m_T^T D W_T) Z - \frac{1}{2} Z^T (K_E+K_T) Z \right) \\
\end{eqnarray}
where
\begin{eqnarray}
W_E & = & \begin{bmatrix} W_S & W_D & 0 \end{bmatrix} \\
W_T & = & \begin{bmatrix} W_S & 0 & W_D \end{bmatrix} \\
K_E & = & \begin{bmatrix} 
   W_S^TDW_S+\frac{1}{2}I & 0 & W_S^T D W_D \\
   0 & 0 & 0 \\   
   W_D^T D W_S & 0 & W_D^T D W_D +  I 
   \end{bmatrix} \\
K_T & = & \begin{bmatrix} 
   W_S^TDW_S+\frac{1}{2}I & W_S^T D W_D & 0 \\
   W_D^T D W_S & W_D^T D W_D + I & 0 \\
   0 & 0 & 0 
   \end{bmatrix}\\
K_E+K_T & = & \begin{bmatrix} 
   2 W_S^TDW_S+I & W_S^T D W_D & W_S^T D W_D \\
   W_D^T D W_S & W_D^T D W_D + I & 0 \\
   W_D^T D W_S & 0 & W_D^T D W_D +  I 
   \end{bmatrix} \\
\end{eqnarray}

We can now find the distribution of $Z$
\begin{eqnarray}
p(Z | M, h_a) & \sim & N(Z | \hat{Z}, \Sigma)
\end{eqnarray}
where
\begin{eqnarray}
\hat{Z} & = & \Sigma \Phi \\
\Phi & = & (W_E^T D m_E + W_T^T D m_T) \\
\Sigma & = & (K_E+K_T)^{-1}
\end{eqnarray}

Finally, evaluating the distribution at 0,
\begin{eqnarray}
\log p(Z = 0| M, h_a) = -\frac{1}{2} \log |\Sigma| -\frac{1}{2} \Phi^T \Sigma \Phi - \frac{N_S+2N_D}{2} \log 2\pi
\end{eqnarray}

To compute the likelihood in Equation (\ref{eq:likM}), we also need the prior of $Z$ evaluated at 0, which is given by
\begin{eqnarray}
\log p(Z = 0| h_a) & = & -\frac{1}{2} \log |\text{diag}(I_{N_S}, I_{N_D}, I_{N_D})| - \frac{N_S+2N_D}{2} \log 2\pi \\
& = &  - \frac{N_S+2N_D}{2} \log 2\pi.
\end{eqnarray}

The needed likelihoods are then given by:
\begin{eqnarray}
\log p(M| h_a) = \frac{1}{2} \log |\Sigma|  + \frac{1}{2} \Phi^T \Sigma \Phi + \log p(M|Z, h_a)
\end{eqnarray}

As mentioned above, the last term is the same for all terms in the LR so it cancels out and does not need to be computed. Defining
\begin{eqnarray}
Q(H_{l_SS},h) = \frac{1}{2} \log |\Sigma| + \frac{1}{2} \Phi^T \Sigma \Phi + \log P(h|H_{l_SS}),
\end{eqnarray}
the log LR (LLR) is then given by
\begin{eqnarray}
LLR & = & \log \sum_{h\in \cal H} \exp Q(H_{SS},h) -  \log \sum_{h\in \cal H} \exp Q(H_{DS},h)
\end{eqnarray}
where the hypotheses determine $P$ and the order in which the matrices $V$ and $U$ are sorted to form $W_S$, $W_D$, two matrices needed to compute $\Sigma$ and $\Phi$.

This equation gives identical results as the equation derived in \cite{ferrer2017joint} for single-enrollment single-test scoring when a single nuisance condition is included in the model.

\bibliographystyle{IEEEtran}

\bibliography{all-short}

\end{document}